\documentclass[letterpaper]{article} % DO NOT CHANGE THIS
\usepackage{aaai2026}  % DO NOT CHANGE THIS
\usepackage{times}  % DO NOT CHANGE THIS
\usepackage{helvet}  % DO NOT CHANGE THIS
\usepackage{courier}  % DO NOT CHANGE THIS
\usepackage[hyphens]{url}  % DO NOT CHANGE THIS
\usepackage{graphicx} % DO NOT CHANGE THIS
\usepackage{natbib}  % DO NOT CHANGE THIS AND DO NOT ADD ANY OPTIONS TO IT
\usepackage{caption} % DO NOT CHANGE THIS AND DO NOT ADD ANY OPTIONS TO IT
\usepackage{algorithm}
\usepackage{algorithmic}

\usepackage{newfloat}
\usepackage{listings}
\usepackage{amssymb} 
\usepackage{amsmath}
\usepackage{booktabs}
\usepackage{multirow}

\DeclareCaptionStyle{ruled}{labelfont=normalfont,labelsep=colon,strut=off} % DO NOT CHANGE THIS
\floatstyle{ruled}
\newfloat{listing}{tb}{lst}{}
\floatname{listing}{Listing}
\title{Negative Entity Suppression for Zero-Shot Captioning with Synthetic Images}
\author{
    Zimao Lu\textsuperscript{\rm 1}, Hui Xu\textsuperscript{\rm 1,\rm 2}\thanks{Corresponding Author}, Bing Liu\textsuperscript{\rm 1,\rm 2}$^*$, Ke Wang\textsuperscript{\rm 1,\rm 2}
}
\affiliations{
    \textsuperscript{\rm 1}School of Computer Science and Technology/ School of Artificial Intelligence,China University of Mining and Technology,  Jiangsu, China\\
    \textsuperscript{\rm 2}Mine Digitization Engineering Research Center of the Ministry of Education\\
    \{luzimao, xuhui, liubing, wangke\}@cumt.edu.cn
}

\usepackage{bibentry}
\begin{document}

\maketitle

\begin{abstract}
Text-only training provides an attractive approach to address data scarcity challenges in zero-shot image captioning (ZIC), avoiding the expense of collecting paired image-text annotations. However, although these approaches perform well within training domains, they suffer from poor cross-domain generalization, often producing hallucinated content when encountering novel visual environments. Retrieval-based methods attempt to mitigate this limitation by leveraging external knowledge, but they can paradoxically exacerbate hallucination when retrieved captions contain entities irrelevant to the inputs. We introduce the concept of \textbf{\emph{negative entities}}—objects that appear in generated caption but are absent from the input—and propose Negative Entity Suppression (NES) to tackle this challenge. NES seamlessly integrates three stages: (1) it employs synthetic images to ensure consistent image-to-text retrieval across both training and inference; (2) it filters negative entities from retrieved content to enhance accuracy; and (3) it applies attention-level suppression using identified negative entities to further minimize the impact of hallucination-prone features. Evaluation across multiple benchmarks demonstrates that NES maintains competitive in-domain performance while improving cross-domain transfer and reducing hallucination rates, achieving new state-of-the-art results in ZIC. Our code is available at https://github.com/nidongpinyinme/NESCap.
\end{abstract}


\begin{figure}[!t]
    \centering
    \includegraphics[width=0.45\textwidth]{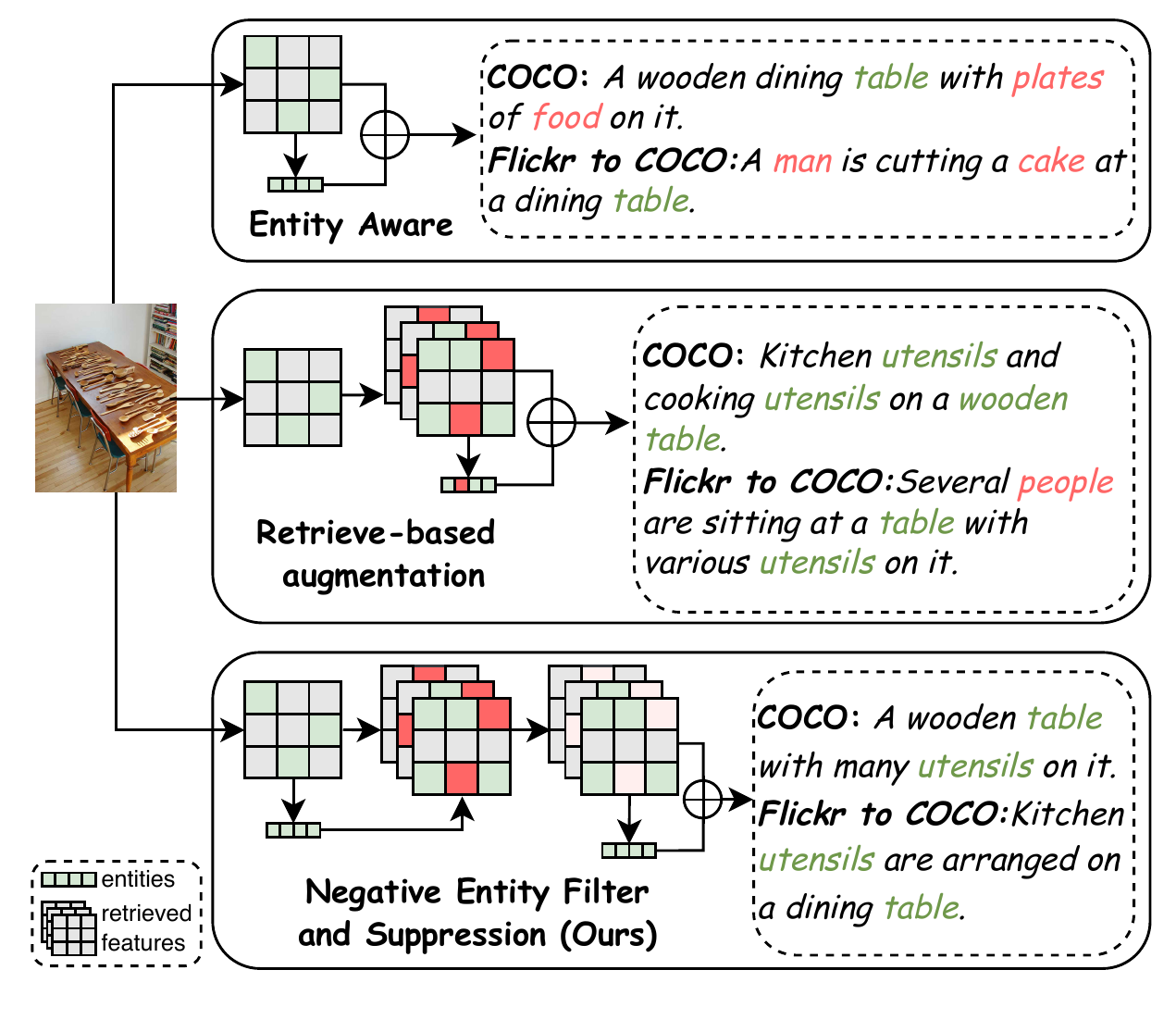}
    \caption{Illustration of cross-domain degradation in ZIC models. COCO: training and inference on COCO dataset; Flickr to COCO: training on Flickr30k and inference on COCO. Correct and incorrect entities are marked in green and red, respectively. Entity-aware models, lacking image-text pairs, tend to generate hallucinated entities with high semantic association to existing entities. Retrieval-based models achieve better in-domain performance but introduce new hallucinations that reduce cross-domain generalization. Our model incorporates identification and suppression modules to effectively mitigate the impact of hallucinated information in retrieved content.}
    \label{fig:intro}
\end{figure}

\section{Introduction}

Image captioning bridges computer vision and natural language processing by generating textual descriptions for visual content using encoder-decoder architectures. While existing approaches achieve impressive results in supervised settings, they rely heavily on large-scale paired text-image annotations, limiting their applicability to out-of-distribution images containing unfamiliar objects. This limitation motivates Zero-shot Image Captioning (ZIC), which aims to generate captions without paired training data, reducing annotation costs and enabling deployment across previously unseen domains. However, as shown in Fig.~\ref{fig:intro}, current ZIC methods face a significant challenge: despite strong in-domain performance, they suffer substantial degradation when applied to cross-domain tasks.

Two interconnected factors contribute to cross-domain performance degradation. First, the \textbf{cross-modal gap}: lacking high-quality visual information during training, models fail to develop robust visual priors and instead over-rely on linguistic patterns learned from text. Second, \textbf{cross-domain hallucination}: when encountering out-of-domain scenarios, models tend to generate objects that are highly associated with existing content but not actually present in the images. These hallucinated entities typically either share visual similarities with actual image content or co-occur with them in training corpora frequently.

\begin{figure}[!t]
    \centering
    \includegraphics[width=0.48\textwidth]{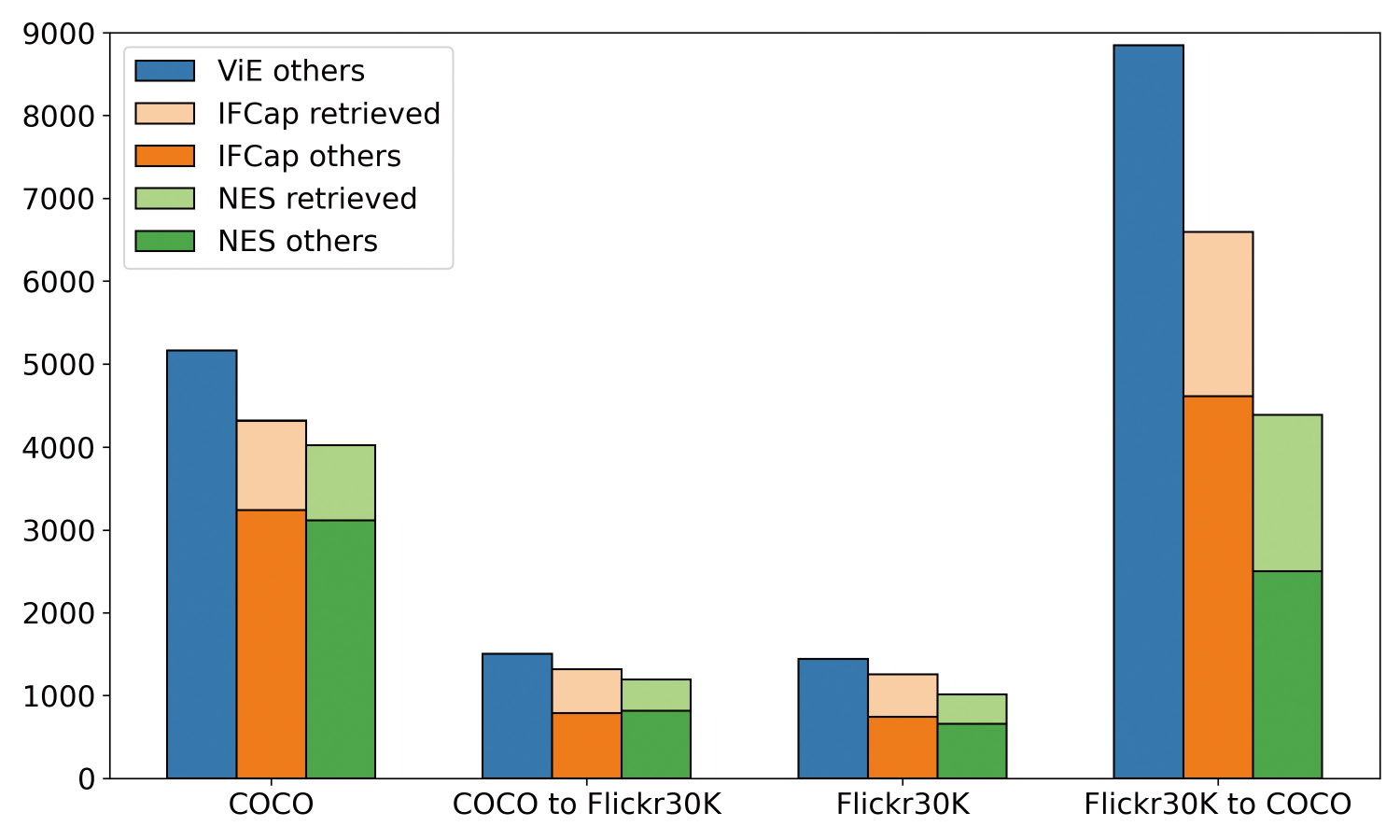}
    \caption{Analysis of hallucination patterns in existing ZIC methods. Hallucinated entities are classified as \textit{retrieved} (originating from retrieved captions) or \textit{others} across four scenarios. The results demonstrate that retrieval-based approaches such as IFCap suffer from the risk of hallucinations, while our NES method effectively reduces both the overall number of hallucinated entities and those specifically induced by retrieval.}
    \label{fig:ana_caps}
\end{figure}

Existing approaches have attempted to address these challenges from various perspectives, such as entity extraction\cite{JunjieFeiTransferableDecoding2023,SoeunLeeIFCapImagelike2024}, data synthesis\cite{LiuImprovingCrossModal2024,MaImageCaptioning2024} and external knowledge enhancement\cite{RamosRetrievalaugmentedImage2023,KimViPCapRetrieval}. However, most methods treat the modality gap and hallucination issues as separate problems, overlooking their intrinsic connection. This motivates a unified framework that can simultaneously mitigate modality gap and suppress hallucinated entities, thereby improving cross-domain generalization in ZIC.

To investigate the underlying causes of cross-domain performance degradation, we analyze hallucination phenomena in existing models. As illustrated in Figure~\ref{fig:ana_caps}, our analysis reveals that while IFCap achieves a lower overall hallucination rate compared to ViE, approximately 20\% of the hallucinated entities are present in the retrieved information(detailed analysis in Appendix~\ref{app:hallucination_analysis}). This finding suggests that identifying and filtering out these entities from the retrieval corpus could potentially improve the model's generation performance. We validate this hypothesis through our proposed NES model, which achieves a 30\% reduction in total hallucinations and a 10\% decrease in retrieval-induced hallucinations on the Flickr30k-to-COCO task.

\begin{figure}[htbp!]
    \centering
    \includegraphics[width=0.48\textwidth]{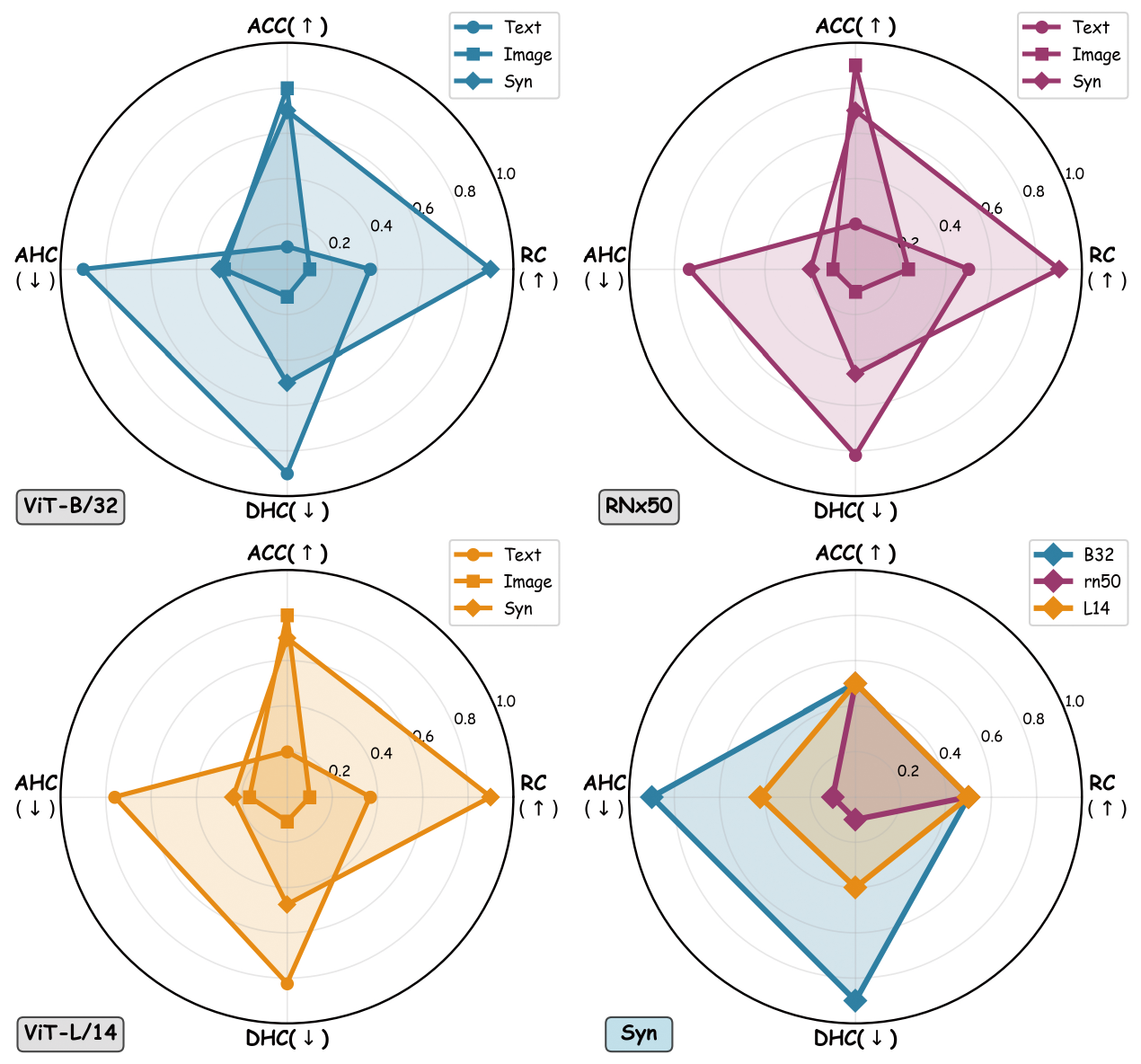}
    \caption{Analysis of retrieval performance across different data sources and CLIP encoders on the COCO dataset (normalized). Higher values indicate better performance for accuracy (ACC) and recall (RC), while lower values indicate better performance for average hallucination count per image (AHC) and deduplicated hallucination count (DHC).}
    \label{fig:intro_retrieval_ana}
\end{figure}

To optimize retrieval-assisted generation while reducing hallucinations, we analyze of the relationships between different data sources, encoders, and retrieval accuracy across the COCO dataset using multiple CLIP encoders.  As illustrated in Figure~\ref{fig:intro_retrieval_ana}, our findings demonstrate that synthetic images achieve superior balance across accuracy and recall metrics while reducing hallucination compared to text-based retrieval, and the RN50 model achieves the lowest hallucination rate while maintaining accuracy and recall on par with other models (detailed analysis in Appendix~\ref{app:retrieval_analysis}).

Motivated by these findings and existing approaches, we propose the \textbf{Negative Entity Suppression (NES)}, a framework that simultaneously addresses modality gap and hallucination challenges through synthetic image-to-text processing and hallucination-aware filtering. Rather than treating these issues separately, NES leverages synthetic images to both enhance text-only training inputs and enable consistent image-to-text retrieval across both training and inference phases. For retrieved entities, we categorize them into positive entities (present in the input) and negative entities (absent from the input), then apply targeted attention-level suppression to reduce the interference of hallucination-prone information in retrieved content. 

In summary, our contributions are as follows:
\begin{itemize}
    \item \textbf{Synthetic Image-to-text Retrieval:} We employ synthetic images generated by diffusion models for both input enhancement and retrieval queries, maintaining consistency between training and inference while reducing modality gap inherent in text-only retrieval approaches.
    \item \textbf{Negative Entity Identification and Filtering:} We develop an entity verification framework that distinguishes between visually grounded and ungrounded entities in retrieved content, utilizing ground-truth captions during training and visual-semantic alignment during inference.
    \item \textbf{Attention-level Hallucination Suppression:} We design a targeted attention mechanism that identifies and suppresses negative entity influences within retrieved caption features, minimizing their impact on feature fusion and overall caption quality.
    \item \textbf{Superior Cross-domain Performance:} We demonstrate that NES maintains in-domain performance while significantly improving cross-domain generalization, achieving a 14\% improvement in CIDEr and reducing hallucination rates by 54\% on the Flickr30k-to-COCO task.
\end{itemize}

\section{Related Work}

\subsection{Text-only Training Image Captioning}
Text-only training approaches aim to enhance the mapping from vision to language in the absence of image–text alignment. A pioneering method, ClipCap~\cite{RonMokadyClipCapCLIP2021}, uses a mapping network to transform visual features into semantic prefixes for language models. Building on this, ViECap~\cite{JunjieFeiTransferableDecoding2023} introduces an entity extraction module that serves as hard prompts for the decoder, mitigating hallucinations caused by overreliance on pretrained knowledge. IFCap~\cite{SoeunLeeIFCapImagelike2024} incorporates a retrieval module to replace image-derived entities with those from retrieved captions for better accuracy. MERCap~\cite{ZengZeroShotImage2025} constructs an image–entity memory bank, retrieving relevant entities during inference to complement extracted entities for caption generation. Diffusion Bridge~\cite{LeeDiffusionBridge2025} leverages diffusion models trained exclusively on text embeddings to progressively align vision embeddings with the text embedding distribution through a reverse diffusion process, effectively reducing the modality gap in CLIP for improved ZIC.

\subsection{Retrieval-based Image Captioning}

Retrieval-based methods in image captioning typically fall into image-to-text and text-to-text retrieval. The former retrieves captions semantically aligned with the input image, while the latter finds similar captions based on a query sentence. These techniques have been widely adopted to enhance captioning quality and transferability. For example, ~\cite{RamosRetrievalaugmentedImage2023} proposes an end-to-end model that jointly embeds images and retrieved captions, while ~\cite{RitaRamosSmallCapLightweight2023} designs a lightweight model that trains only cross-attention layers for better generalization across domains.

In ZIC, retrieval modules are commonly designed as plug-and-play components for domain adaptation. ~\cite{ZequnZengMeaCapMemoryAugmented2024} employs image-to-text retrieval from a large-scale textual memory bank through semantic similarity matching and introduces a retrieve-then-filter mechanism to extract and refine the most relevant entity prompts for caption generation; ~\cite{SoeunLeeIFCapImagelike2024} employs a dual-phase retrieval strategy that performs text-to-text retrieval during training to align textual features and switches to image-to-text retrieval during inference; ~\cite{YangReViLMRetrievalAugmented2023a} adopts image-to-image retrieval based on visual feature similarity to retrieve the most visually analogous images from a reference database and leverages their corresponding captions as retrieval-augmented guidance for zero-shot generation. 

\begin{figure*}[!t]
    \centering
    \includegraphics[width=0.95\textwidth]{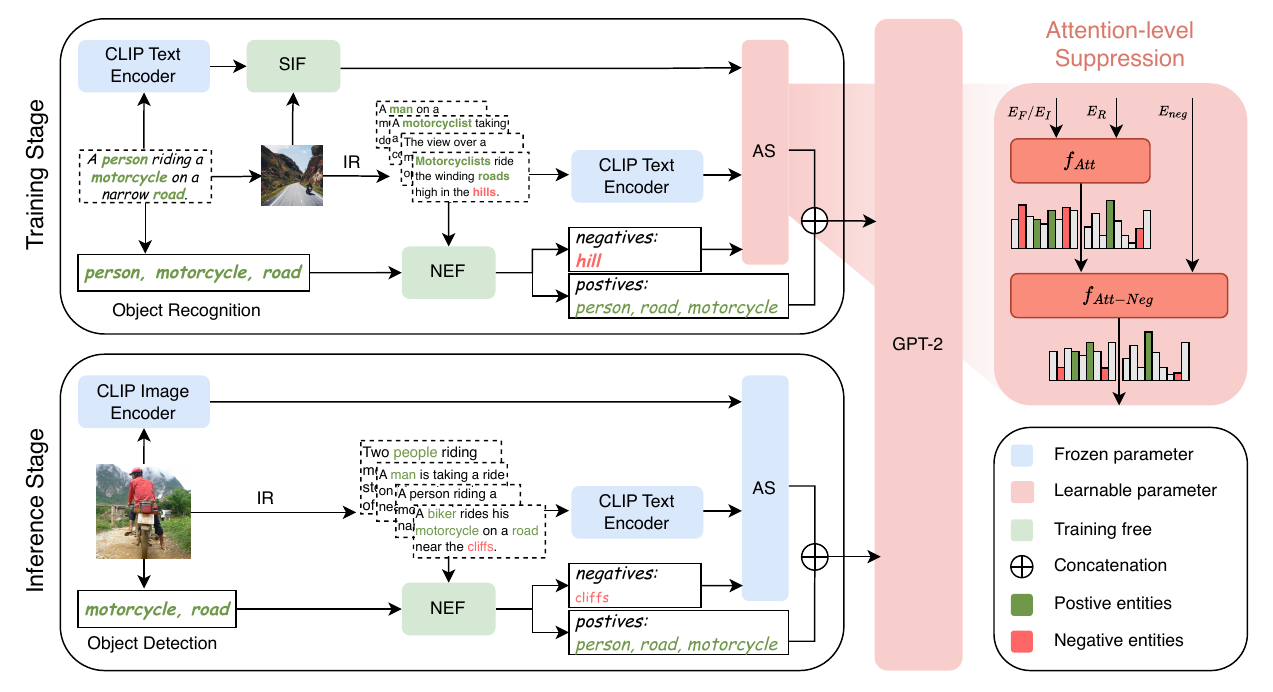}
    
    \caption{Overall framework of NES. (a) Training phase: The framework generates synthetic images from input text using diffusion models, then retrieves relevant captions via IR (Image-to-text Retrieval). The SIF (Synthetic Image Fusion) module enhances input text features using synthetic images. The NEF (Negative Entity Filtering) module categorizes retrieved entities into positive and negative sets based on input text entities. The AS (Attention-level Suppression) module first fuses retrieved captions with enhanced input features, then suppresses hallucination-prone features using negative entities. Final features are concatenated with positive entities and fed to GPT-2 for generation. (b) Inference phase: The pipeline directly uses input images for retrieval without SIF enhancement. NEF filtering is performed using image-extracted entities with CLIP similarity. The AS module remains consistent with training.}
    \label{fig:framework}
\end{figure*}

\subsection{Hallucination Suppression in Image Captioning}
Hallucination has become a prominent issue in image captioning, especially in zero-shot or out-of-distribution settings. Hallucinations occur when the model generates objects, attributes, or relations not present in the image, undermining the model’s reliability.

To quantify hallucination, \cite{AnnaRohrbachObjectHallucination2018} proposed the CHAIR metric, which identifies unmatched entities in generated captions. \cite{PetrykALOHaNew2024} extended this with ALOHa to handle open-vocabulary hallucinations. With those metrics, \cite{SarkarMitigatingHallucinations2025} demonstrated that imbalanced cross-modal attention distribution causes models to over-rely on language priors, and proposed suppressing non-visual attention heads. \cite{HuangOPERAAlleviating2024a} found that summary tokens from self-attention mislead generation and proposed OPERA to selectively downweight these misleading summary tokens during generation. \cite{LengMitigatingObject2023} identified hallucinated entities by comparing the model’s output on perturbed and original images.

\section{Methodology}

Text-only training methods face two challenges: the cross-modal gap between textual training data and visual inference inputs, and the cross-domain hallucinations when encountering out-of-domain scenarios. These challenges are particularly pronounced in zero-shot scenarios, where models must generate captions for unseen visual domains without corresponding training examples.

To address these challenges, we employ synthetic images to enable image-to-text retrieval during training and suppress hallucination-prone features using negative entities filtered from retrieved captions. Our framework integrates three components: (1) \textbf{Image-to-text Retrieval} for modality consistency, (2) \textbf{Negative Entity Filtering} to improve generation accuracy, and (3) \textbf{Attention-level Suppression} to reduce hallucination tendency.

\subsection{Image-to-text Retrieval}

Text-to-text retrieval introduces hallucination information that degrades performance, as text-based queries lack visual grounding and result in retrieval of semantically similar but visually irrelevant content. As demonstrated in our analysis mentioned before, synthetic images reduce hallucination rates while maintaining retrieval quality. We therefore use synthetic images as retrieval queries to bridge the modality gap and ensure consistent image-to-text retrieval across training and inference.

\noindent\textbf{Synthetic Image Generation.} For ground-truth captions $T$, we employ a pretrained diffusion model to generate corresponding synthetic images $\tilde{I}$. These synthetic images serve as visual queries for retrieval, enabling visual-semantic alignment capture while reducing dependency on textual priors. Since these captions are not designed as generation prompts and diffusion models may introduce biases, the resulting images may not perfectly capture the original descriptions. We quantify this limitation in Table~\ref{tab:synthetic_quality} and address it through quality-based filtering and multimodal fusion.

\noindent\textbf{Synthetic Image Filtering and Fusion.} To address potential quality issues with synthetic images, we compute synthetic image embedding $\mathbf{E}_{\tilde{I}} = \text{CLIP}_v(\tilde{I})$ for synthetic images $\tilde{I}$ and text embedding $\mathbf{E}_T = \text{CLIP}_t(T)$ for ground-truth captions $T$, then calculating the CLIPScore as:
\begin{equation}
\text{CLIPScore}(\mathbf{E}_{\tilde{I}}, \mathbf{E}_T) = \frac{\mathbf{E}_{\tilde{I}}  \cdot \mathbf{E}_T}{{||\mathbf{E}_{\tilde{I}} }||\cdot| |\mathbf{E}_T||}
\end{equation}
We store CLIPScores for all synthetic image-text pairs and discard those below threshold $\tau$ to ensure quality.

To enhance representation quality, we combine synthetic image features with their corresponding text embeddings. Since synthetic images may not capture all semantic nuances while text embeddings lack visual grounding, fusing both modalities creates more robust representations. 

\begin{table*}[!ht]
    \centering
    {
    \begin{tabular}{l|c|c|cccc|ccccc}
        \toprule
        \multirow{2}{*}{\textbf{Method}} & \multirow{2}{*}{\textbf{Encoder}} & \multirow{2}{*}{\textbf{Decoder}} & \multicolumn{4}{c}{\textbf{COCO}} & \multicolumn{4}{|c}{\textbf{Flickr30k}} \\
        \cmidrule(lr){4-7} \cmidrule(lr){8-11}
        & & & B@4 & M & C & S & B@4 & M & C & S  \\
        
         \midrule
        CapDec (2022)  &  RN50 $\times$ 4 & GPT-2\textsubscript{Large} & 26.4 & 25.1 & 91.8 & 11.9 & 17.7 & 20.0 & 39.1 & 9.9 \\
        DeCap (2023) & ViT-B/32 & Transformer\textsubscript{Base} & 24.7 & 25.0 & 91.2 & 18.7 & 21.2 & 21.8 & 56.7 & 15.2 \\
        ViECap (2023) & ViT-B/32 & GPT-2\textsubscript{Base} & 27.2 & 24.8 & 92.9 & 18.2 & 20.3 & 20.2 & 47.8 & 13.6 \\
        SynTIC(2023) & ViT-B/32 & Transformer$_{H=4}^{L=4}$ & \underline{29.9} & 25.8 & 101.1 & 19.3 & 22.3 & \underline{22.4} & 56.6 & \underline{16.6} \\
        ICSD(2023) & ViT-B/32 & BERT\textsubscript{Base} & \underline{29.9} & 25.4 &  96.6 & - & \textbf{25.2} & 20.6 & 54.3 & -\\
        IFCap (2024) & ViT-B/32 & GPT-2\textsubscript{Base} & \textbf{30.8} & \underline{26.7} & \underline{108.0} & \underline{20.3} & 23.2 & \textbf{22.9} & \underline{64.4} & \textbf{17.0} \\
        \midrule
        NES & ViT-B/32 & GPT-2\textsubscript{Base} & \textbf{30.8} & \textbf{26.8} & \textbf{109.9} & \textbf{20.6} & \underline{24.3} & 22.1 & \textbf{66.8} & 15.9 \\
        \bottomrule
    \end{tabular}}
    \caption{The results of the in-domain captioning included the COCO test split and the Flickr30k test split.}
    \label{tab:indomain}
\end{table*}

\noindent\textbf{Image-to-text Retrieval and Mapping.} For retrieval, we use filtered synthetic images during training and input images during inference, the detailed retrieval process is described in Appendix~\ref{app:image-to-text retrieval}. We integrate retrieved captions through a multi-step feature fusion process. First, we encode the retrieved captions $\mathbf{E}_R = \mathrm{CLIP}_t(R)$ and adjust their feature dimensions via linear projection layers $f_{l1}$ and $f_{l2}$. The cross-attention module $f_{\mathrm{Att}}$ then fuses the processed retrieval features with the fused synthetic image-text features to obtain $\mathbf{E}_{\mathrm{Attn}}$. Finally, $\mathbf{E}_{\mathrm{Attn}}$ is fed into a trainable mapping network $f_{\mathrm{Map}}$ that encodes the semantic representation of the input. This process can be formalized as:

\begin{align}
    \mathbf{E}_{\mathrm{Attn}} &= f_{\mathrm{Att}}\big(f_{l1}(\mathbf{E}_{\mathrm{F}}),\, f_{l2}(\mathbf{E}_R)\big) \\
    \mathbf{E}_{\mathrm{map}} &= f_{\mathrm{Map}}(\mathbf{E}_{\mathrm{Attn}})
\end{align}

\subsection{Negative Entity Filtering}
Retrieved captions often contain entities that are semantically inconsistent with the input image, particularly in cross-domain scenarios where domain shift exacerbates this mismatch. Such inconsistent entities can introduce hallucinations that mislead the caption generation process. To address this problem, we propose entity-aware processing that explicitly identifies negative entities.

\noindent\textbf{Entity Identification During Training.} During training, we leverage the availability of ground-truth captions to establish supervised entity classification. We extract key entities $e_{\text{key}}$ from ground-truth captions and candidate entities $e_{\text{can}}$ from retrieved captions using parsing tools. Key entities automatically become positive entities $e_{\text{pos}}$ (i.e., $e_{\text{pos}} = e_{\text{key}}$), while candidate entities not appearing in $e_{\text{key}}$ are labeled as negative entities $e_{\text{neg}} = e_{\text{can}} \setminus e_{\text{key}}$.

\noindent\textbf{Entity Identification During Inference.} During inference, ground-truth captions are unavailable, making it challenging to directly identify positive entities $e_{\text{pos}}$. We design a similarity-based filtering mechanism that leverages visual-semantic alignment for reliable entity selection.

The filtering process operates as follows: We use the ViECap method to extract key entities $e_{\text{key}}$ from the input image. The detailed extraction process is provided in the Appendix~\ref{app:Clip-based Entity Classification}. From retrieved captions, we extract candidate entities $e_{\text{can}}$. For each candidate entity not in $e_{\text{key}}$, we compute its similarity with the image representation. Entities exceeding threshold $\tau_{sim}$ are added to positive entities $e_{\text{pos}}$, while others become negative entities $e_{\text{neg}}$. This process is formulated as:
\begin{align}
e_{\text{filter}} &= e_{\text{can}} \setminus e_{\text{key}}\\
e_{\text{pos}} &= e_{\text{key}} \cup \left\{ e \in e_{\text{filter}} \mid S_{\text{sim}}(e,E_I) > \tau_{sim} \right\} \\
e_{\text{neg}} &= e_{\text{filter}} \setminus e_{\text{pos}}
\end{align}
where $S_{\text{sim}}(e,E_I)$ is the cosine similarity function.

\subsection{Attention-level Hallucination Suppression}
To further leverage negative entities to improve the accuracy of generated captions and reduce hallucination tendencies caused by retrieved captions and language priors, we identify and suppress features related to negative entities in the fused features, thereby further enhancing the model's generalization capability.

\noindent\textbf{Hallucination-prone Feature Filtering.} To selectively suppress features associated with negative entities, we first encode all negative entities $\mathbf{e}_{\text{neg}}$ using the CLIP text encoder and then apply a cross-attention mechanism $f_{\text{neg}}$ to compute attention weights between the negative entity embeddings $\mathbf{E}_{\text{neg}}$ (as queries) and the fused feature representations $\mathbf{E}_{\text{F}}$ (as keys and values). This process identifies which feature dimensions exhibit strong correlated with hallucination-prone entities. Based on the computed attention weights, we employ threshold-based filtering strategies to suppress the contribution of high-correlation features, thereby reducing hallucination tendencies in the final representations. Different filtering strategies are evaluated in the ablation study.

\noindent\textbf{Hallucination-prone Feature Suppression.} Based on the attention weights computed in the filtering stage, we identify features that exhibit strong correlations with negative entities. For features whose attention weights exceed threshold $\tau_{\text{neg}}$, we apply a suppression factor $\lambda < 1$ to reduce their contribution to the decoder during caption generation. This selective suppression mechanism is formulated as:

\begin{align}
\mathbf{E}_{\text{A}} &= f_{\text{neg}}(\mathbf{E}_{\text{neg}},\mathbf{E}_{\mathrm{map}}) \\
\mathbf{E}_{\text{sup}} &=
\begin{cases}
\lambda \cdot \mathbf{E}_{\text{map}}^{(h)}, & \text{if }  \mathbf{E}_{\text{A}}^{(h)} > \tau_{\text{neg}} \\
\mathbf{E}_{\text{map}}^{(h)}, & \text{otherwise}
\end{cases}
\end{align}
where $\mathbf{E}_{\text{sup}}$ represents the suppressed feature representations, $\tau_{\text{neg}}$ is the threshold of hallucinations, and $\lambda$ is a hyperparameter that controls the suppression strength.

\section{Experiments}

\subsection{Experimental Details}
\noindent\textbf{Implementation Details.} We employ Stable Diffusion v1.5 for synthetic image generation at 512$\times$512 resolution with 20 denoising steps. CLIP-ViT-B/32 serves as both the image and text encoder, while GPT-2$_\text{base}$ acts as the text decoder. During training, we freeze the encoder parameters and only update the AS Module and text decoder. The model is trained for 5 epochs with a batch size of 80 using the AdamW optimizer with an initial learning rate of $2 \times 10^{-5}$ and cosine annealing scheduling. For threshold selection, we set synthetic image filtering threshold $\tau = 0.6$, entity similarity threshold $\tau_{sim} = 0.2$, suppression strength $\lambda$ is set to 0.3. All experiments are conducted on a single NVIDIA RTX 4080 Super GPU with 16GB VRAM, requiring approximately one hour and 12GB of GPU memory for training.

\noindent\textbf{Datasets and Metrics.} For in-domain evaluation, we conduct experiments on the MS-COCO~\cite{ChenMicrosoftCOCO2015} and Flickr30k~\cite{PeterYoungimagedescriptions2014} datasets using the Karpathy split~\cite{AndrejKarpathyDeepVisualSemantic2015}. For cross-domain evaluation, we assess performance on the NoCaps~\cite{Agrawalnocapsnovel2019} validation set, which contains novel object categories not seen during training. We adopt standard image captioning metrics including CIDEr~\cite{RamakrishnaVedantamCIDErConsensusbased2015}, BLEU@4~\cite{KishorePapineniBleuMethod2002}, METEOR~\cite{BanerjeeMETEORAutomatic2005}, and SPICE~\cite{AndersonSPICESemantic2016}. To evaluate hallucination suppression effectiveness, we report CHAIR~\cite{RohrbachObjectHallucination2019} scores on COCO dataset, which measure the proportion of objects that are not present in the ground-truth captions. We also compute entity recall rates to assess how well our method captures ground-truth entities in the generated captions.

\noindent\textbf{Baselines.} We compare our model with several state-of-the-art text-only image captioning methods. CapDec~\cite{DavidNukraiTextOnlyTraining2022} and ViECap~\cite{JunjieFeiTransferableDecoding2023} build upon ClipCap~\cite{RonMokadyClipCapCLIP2021}, using Gaussian noise to align text and image features, with ViECap additionally incorporating entity extraction for hallucination mitigation. CLOSE~\cite{SophiaGuCantBelieve2023} explores various noise configurations, while DeCap~\cite{WeiLiDeCapDecoding2023} introduces a memory bank mechanism for improved retrieval. IFCap~\cite{SoeunLeeIFCapImagelike2024} employs retrieval-based entity filtering with frequency-based selection to replace image-derived entities with retrieved ones for enhanced accuracy. ICSD~\cite{MaImageCaptioning2024} and SynTIC~\cite{LiuImprovingCrossModal2024} represent recent advances that leverage text-to-image generation models like Stable Diffusion~\cite{RombachHighResolutionImage2022} to bridge the modality gap between text-only training and visual inference.
\subsection{In-domain captioning}
We evaluate NES on in-domain settings using the COCO and Flickr30k datasets, with results reported in Table~\ref{tab:indomain}. Compared with previous state-of-the-art text-only image captioning methods, NES achieves superior performance across all metrics on the COCO dataset, even outperforming models that employ larger architectures. On Flickr30k, NES demonstrates competitive performance on BLEU@4 and METEOR while achieving the highest CIDEr score, highlighting its effectiveness in generating accurate and contextually relevant captions.

\subsection{Cross-domain captioning}

We assess the transferability of NES across diverse domains through cross-domain scenarios between COCO and Flickr30k datasets, as well as the NoCaps validation set, with results presented in Table~\ref{tab:crossdomain} and Table~\ref{tab:nocaps}.

In cross-domain settings between COCO and Flickr30k, NES achieves state-of-the-art performance across most metrics in both transfer directions. Particularly noteworthy is the Flickr30k-to-COCO transfer scenario, we provide results for both inference with and without target-domain retrieval texts, where NES achieves substantial improvements with CIDEr scores of 69.9 compared to the best baseline IFCap's 60.7.
\begin{table}[htbp]
    \centering
    \scalebox{0.77}{
    \begin{tabular}{l|cccc|cccc}
    \toprule
    \multirow{2}{*}{\textbf{Methods}} & \multicolumn{4}{c|}{\textbf{MSCOCO} $\rightarrow$ \textbf{Flickr30k}} & \multicolumn{4}{c}{\textbf{Flickr30k} $\rightarrow$ \textbf{MSCOCO}} \\
    \cmidrule(lr){2-5} \cmidrule(lr){6-9}
    & B@4 & M & C & S & B@4 & M & C & S \\
    \midrule
    \multicolumn{9}{l}{\textbf{Without target domain’s corpus}} \\
    \midrule
    DeCap & 16.3 & 17.9 & 35.7 & 11.1 & 12.1 & 18.0 & 44.4 & 10.9\\
    ViECap & 17.4 & 18.0 & 38.4 & 11.2 & 12.9 & 19.3 & 54.2 & 12.5\\
    % Knight & \textbf{21.1} & \textbf{22.0} & 48.9 & 14.2 & \textbf{19.0} & 22.8 & 64.4 & 15.1\\
    SynTIC & 17.9 & 18.6 & 38.4 & 11.9 & 14.6 & 19.4 & 47.0 & 11.9\\
    IFCap & 17.8 & 19.4 & 47.5 & 12.7 & 14.7 & 20.4 & 60.7 & 13.6\\
    \midrule
    NES & \textbf{18.8} & \textbf{20.0} & \textbf{52.3} & \textbf{13.7} &  \textbf{17.3} & \textbf{20.7} & \textbf{69.9} & \textbf{15.0}\\
    \midrule
    \multicolumn{9}{l}{\textbf{With target domain’s corpus}} \\
    \midrule
    SynTIC$-TT$ & 19.4 & 20.2 & 43.2 & 13.9 & 20.6 & 21.3 & 64.4 & 14.3\\
    IFCap$-TT$ & 21.2 & \textbf{21.8} & 59.2 & \textbf{15.6} & 19.0 & 23.0 & 76.3 & 17.3\\
    \midrule
    NES$-TT$ & \textbf{21.3} & 21.4 & \textbf{59.3} & 14.9 & \textbf{22.9} & \textbf{23.8} & \textbf{87.5} & \textbf{18.6} \\
    \bottomrule
    \end{tabular}}
    \caption{Cross-Domain Evaluation. X $\rightarrow$ Y means source domain to target domain. $-TT$ : models can access to target domain’s corpus during inference time.}
    \label{tab:crossdomain}
\end{table}

\begin{table}[htbp]
    \centering
    \scalebox{0.77}{
    \begin{tabular}{l|cc|cc|cc|cc}
    \toprule
    \multirow{2}{*}{\textbf{Methods}}  & \multicolumn{2}{|c|}{\textbf{in-domain}} & \multicolumn{2}{|c|}{\textbf{near-domain}} & \multicolumn{2}{|c|}{\textbf{out-of-domain}} & \multicolumn{2}{|c}{\textbf{Overall}} 
    \\ \cmidrule{2-3} \cmidrule{4-5} \cmidrule{6-7} \cmidrule{8-9}
    & C & S & C & S & C & S & C & S \\
    \midrule
    CapDec & 60.1 & 10.2 & 50.2 & 9.3 & 28.7 & 6.0 & 45.9 & 8.3\\
    DeCap & 65.2 & - & 47.8 & - & 25.8 & - & 45.9 & - \\
    ViECap & 61.1 & 10.4 & 64.3 & 9.9 & 65.0 & 8.6 & 66.2 & 9.5 \\
    IFCap & \textbf{75.8} & \textbf{12.4} & 72.3 & 11.6 & 60.2 & 8.9 & 70.5 & 10.8\\
    IFCap$^{\star}$ & 70.1 & 11.2 & 72.5 & 10.9 & 72.1 & 9.6 & 74.0 & 10.5 \\
    \midrule
    NES & 74.7 & 11.7 & \textbf{75.4} & \textbf{11.7} & \textbf{72.6} & \textbf{10.2} & \textbf{76.2} & \textbf{11.2} \\
    % IFCap \cite{} & ViT-B/32 + GPT-2\textsubscript{Base} & 70.1 & 11.2 & \textbf{72.4} & \textbf{10.9} & \textbf{72.1} & 9.6 & \textbf{74.0} & \textbf{10.5} \\
    \bottomrule
        \end{tabular}}
    \caption{Cross-domain captioning results on the NoCaps validation set. $\star$:without Entity Filtering module in the inference time.}
    \label{tab:nocaps}
\end{table}
\subsection{Hallucination Suppression Analysis}
To evaluate the effectiveness of our negative entity suppression mechanism, we analyze hallucination rates using the CHAIR metric on COCO in-domain and Flickr30k-to-COCO cross-domain scenarios, as shown in Table~\ref{tab:chair}. CHAIR measures the proportion of generated objects that are not present in the ground-truth captions, providing a direct assessment of hallucination suppression capability. Specifically, CHAIR-S (sentence-level) measures the percentage of generated captions containing at least one hallucinated object, while CHAIR-I (instance-level) measures the percentage of hallucinated objects among all generated objects. 

NES achieves lower hallucination rates than baseline methods in both evaluation settings. The improvement is more notable in cross-domain scenarios, where baseline methods experience performance drops while NES maintains stable hallucination rates. This confirms that our negative entity suppression approach effectively reduces hallucinated content generation across different domains.

\begin{table}[htbp]
    \centering
    \scalebox{1}{
    \begin{tabular}{l|ccc|ccc}
    \toprule
    \multirow{2}{*}{\textbf{Method}} & \multicolumn{3}{c|}{\textbf{COCO In-domain}} & \multicolumn{3}{c}{\textbf{Flickr30k→COCO}} \\
    \cmidrule(lr){2-4} \cmidrule(lr){5-7}
    & C-S & C-I &Recall& C-S & C-I & Recall\\
    \midrule
    % Add baseline and NES results here
    ViECap & 16.2 & 10.8 &  \textbf{43.7}& 47.2 & 27.3 & \textbf{45.9}\\
    IFcap & 9.7 & 6.6  & 42.9& 22.8 & 15.1 & 42.3 \\
    NES & \textbf{\textit{8.1}} & \textbf{\textit{5.6}}  & 43.2& \textbf{\textit{12.1}} & \textbf{\textit{8.0} }& 44.9\\
    \bottomrule
    \end{tabular}}
    \caption{CHAIR scores and entity recall on COCO in-domain and Flickr30k-to-COCO cross-domain scenarios.\textbf{ \textit{Lower CHAIR}} values indicate better hallucination suppression. \textbf{Higher recall} values indicate better coverage of ground-truth entities.}
    \label{tab:chair}
\end{table}

\subsection{Ablation Study}
We conduct extensive experiments to identify the impact of each key component in NES: Synthetic Images Retrieval (SIR), Synthetic Images Fusion (SIF), Negative Entity Filtering (NEF), and Attention-level Suppression (AS). Given that our primarily aims to improve cross-domain generalization, we conduct ablation studies on the challenging Flickr30k-to-COCO transfer scenario to better demonstrate the effectiveness of each component in cross-domain settings.

\begin{table}[htbp]
    \centering
    \scalebox{0.82}{
    \begin{tabular}{cccc|ccccc}
        \toprule
        \textbf{SIR} & \textbf{SIF} & \textbf{NEF} & \textbf{AS} & \textbf{B@4} & \textbf{CIDEr} & \textbf{C-S} & \textbf{C-I} & \textbf{Recall}\\
        \midrule
        \checkmark & \checkmark & \checkmark & \checkmark & \textbf{22.9} & \textbf{87.5} & 12.1 & 8.0 & 44.9 \\
         & \checkmark & \checkmark & \checkmark & 20.0 & 79.9 & 16.4 & 10.3 & 44.5 \\
        \checkmark & & \checkmark & \checkmark & 20.7 & 82.9 & 12.0 & 7.9  & 44.7\\
        \checkmark & \checkmark & \checkmark &  & 21.4 & 83.9 & 11.8 & 7.9 & 44.1 \\
        \checkmark & \checkmark & &  & 21.4 & 83.0 & 8.6 & \textit{\textbf{6.2}} & 40.8 \\
        \checkmark & & & & 21.0 & 83.3 & \textit{\textbf{8.4}} &  \textit{\textbf{6.2}} & 41.4\\
         & \checkmark & & & 13.8 & 50.2 & 21.8 & 18.6 & 31.9 \\
        &  & \checkmark & \checkmark & 19.1 & 77.7 & 23.9 & 14.3 & \textbf{45.3} \\
        & & & & 19.0 & 76.3 & 22.8 & 15.1 &42.3\\
        \bottomrule
    \end{tabular}}
    \caption{Ablation studies of the key components of NES.}
    \label{tab:ablation}
\end{table}

\noindent\textbf{Ablation Settings:} To systematically analyze each component's contribution in cross-domain scenarios, we conduct ablation studies on the Flickr30k-to-COCO task. For the Synthetic Image-based Retrieval (SIR) module, we replace synthetic images with original text captions as retrieval queries while maintaining the identical retrieval and processing pipeline. For Synthetic Image Fusion (SIF), we remove the fusion mechanism and use only text features for generation. For Negative Entity Filtering (NEF), we disable entity filtering by skipping the negative entity identification step, allowing all retrieved entities to pass through the pipeline. For Attention-level Suppression (AS), we disable the attention-based suppression mechanism. The results are shown in Table~\ref{tab:ablation}.

\noindent\textbf{Synthetic Image-based Retrieval:} When using SIR, all metrics except recall show significant improvements, especially hallucinations decrease about 50\%. This indicates that synthetic image-based retrieval identifies more semantically relevant captions compared to text-based methods. Notably, SIR alone achieves the lowest hallucination rates, confirming its effectiveness in reducing false content. To further examine SIR's impact, we vary the number of retrieved descriptions. Table~\ref{tab:retrieval_num} shows performance degradation with excessive retrievals. While recall shows an increasing trend as the number of retrieved captions grows, analysis reveals that numerous retrieval captions introduce synonym confusion (e.g., `tv' vs `television').

\noindent\textbf{Synthetic Image Fusion:} SIF is a simple yet important module designed to combine the richness of visual features with the accuracy of textual descriptions for optimal decoding performance. To achieve the best results, we explore different fusion strategies as shown in Table~\ref{tab:fusion_strategy}. When using CLIPScore $\alpha$ for fusion, we test both forward fusion ($\alpha \cdot E_{\tilde{I}} + (1-\alpha) \cdot E_T$) and reverse fusion ($(1-\alpha) \cdot E_{\tilde{I}} + \alpha \cdot E_T$). Additionally, we evaluate the effects of different fixed values for fusion weights. The results reveal an interesting trade-off: CLIPScore-based fusion achieves higher captioning scores and recall rates, while fixed values can achieve lower hallucination rates. This suggests that dynamic weighting based on CLIPScore optimizes overall caption quality, whereas fixed weighting provides more stable suppression of hallucinated content. 

\noindent\textbf{Negative Entity Filtering:} NEF serves as the most critical component of our framework, designed to identify and filter out entities absent from images while preserving generation quality. To evaluate NEF's effectiveness, we calculate the number of hallucinated entities in generated captions and their sources. Hallucinations are categorized as retrieval-sourced if they appear in retrieved entities, or model-sourced otherwise. Table~\ref{tab:nef_analysis} compares three filtering strategies: GT (ground truth filtering), CLIP-Det (CLIP detection filtering), and our NEF approach. The results demonstrate that NEF effectively reduces retrieval-sourced hallucinations while maintaining competitive generation quality, showing that our approach strikes an optimal balance between filtering precision and content preservation. 

\noindent\textbf{Attention-level Suppression:} Although we filter out negative entities from retrieved entity lists, these negative entities still exist within the retrieved captions used for feature fusion. AS performs feature-level fine-grained suppression by extracting negative entity-related categories from the fused features. This completes the final piece of our hallucination suppression framework, achieving a comprehensive pipeline from hallucination identification to suppression, which enables the model to achieve outstanding performance in cross-domain and generalization tasks.

To validate the effectiveness of our attention-level suppression mechanism, we conduct two additional ablation studies examining different suppression strategies and intensity levels. Table~\ref{tab:suppression_strategy} compares various approaches for identifying attention heads to suppress, while Table~\ref{tab:suppression_strength} analyzes the impact of different suppression intensities.

The results in Table~\ref{tab:suppression_strength} reveals that moderate suppression ($\lambda = 0.3$) achieves optimal performance, striking a balance between reducing negative entity influence and preserving useful attention patterns. Excessive suppression ($\lambda = 0.0$) degrades performance by removing beneficial attention information, while insufficient suppression ($\lambda = 0.7, 1.0$) fails to adequately mitigate hallucinations.

\section{Conclusion}

This paper presents NES, a framework that addresses cross-
domain degradation in ZIC. The core insight is that text-only training creates inconsistencies with visual inference, leading to hallucinations that harm cross-domain performance. Our solution employs synthetic images across training for consistent retrieval, filters negative entities from retrieved content, and applies attention-level suppression to reduce their influence in fused features. Experiments demonstrate improvements over existing methods, with notable improvements in cross-domain scenarios. Our work validates two key principles: synthetic images can effectively bridge modality gaps in cross-modal learning, and targeted hallucination suppression enables robust cross-domain generalization in zero-shot captioning tasks. Future work will explore advanced strategies for handling semantically similar negative entities and developing more precise suppression mechanisms.

\bibliography{aaai2026}

\clearpage

\appendix
\setcounter{secnumdepth}{1} % Enable section numbering for appendix

\section{Ablation Study}
% \section{Ablation Study on Retrieval Number}
\begin{table}[h]
    \centering
    \scalebox{0.95}{
    \begin{tabular}{c|ccccc}
        \toprule
        \textbf{Retrieval Num} & \textbf{B@4} & \textbf{CIDEr} & \textbf{C-S} & \textbf{C-I} & \textbf{Recall} \\
        \midrule
        9 & \textbf{22.9} & \textbf{87.5} & \textit{\textbf{12.1}} & \textit{\textbf{8.0}} & 44.9  \\
        15 & 21.2 & 85.1 & 12.6 & 8.3 & 45.0  \\
        20 & 19.6 & 80.6 & 13.6 & 8.9 & 45.5 \\
        50 & 12.9 & 49.6 & 20.4 & 12.5 & \textbf{46.4 }\\
        100 & 8.3 & 21.6 & 25.3 & 16.2 & 45.0  \\
        \bottomrule
    \end{tabular}}
    \caption{Ablation study on retrieval number in Flickr30k-to-COCO transfer scenario.}
    \label{tab:retrieval_num}
\end{table}

% \section{Ablation Study on Fusion Strategies}

\begin{table}[h]
    \centering
    \scalebox{0.85}{
    \begin{tabular}{l|ccccc}
        \toprule
        \textbf{Fusion Strategy} & \textbf{B@4} & \textbf{CIDEr} & \textbf{C-S} & \textbf{C-I} & \textbf{Recall}\\
        \midrule
        CLIPScore (Forward) & \textbf{22.9} & \textbf{87.5}  & 12.1& 8.0 & \textbf{44.9}  \\
        CLIPScore (Reverse) & 22.0 & 84.5 & 12.9 & 8.6 & 44.4 \\
        Image only($\alpha=1$) & 21.4 & 83.8 & 11.7 & 7.9& 44.0 \\
        Fixed $\alpha=0.9$ & 21.9 & 84.3 & \textit{\textbf{11.6}} & \textit{\textbf{7.6}} & 44.0  \\
        Fixed $\alpha=0.5$ & 21.2 & 83.4 & 12.0 & 7.8 & 44.3  \\
        Text only($\alpha=0$) & 20.0 & 79.9 & 16.4 & 10.3 & 44.5  \\
        % Fixed $\alpha=0.3$ & 22.5 & 87.3 & 7.2 & 5.3 & 41.1  \\
        % Fixed $\alpha=0.9$ & 22.4 & 86.9 & 7.1 & 5.4 & 40.8 \\
        \bottomrule
    \end{tabular}}
    \caption{Ablation study on fusion strategies in SIF module. Forward: $\alpha \cdot E_{\tilde{I}} + (1-\alpha) \cdot E_T$; Reverse: $(1-\alpha) \cdot E_{\tilde{I}} + \alpha \cdot E_T$.}
    \label{tab:fusion_strategy}
\end{table}

% \section{Ablation Study on Selection Strategies}
\begin{table}[h]
    \centering
    \scalebox{0.85}{
    \begin{tabular}{l|ccccc}
        \toprule
        \textbf{Suppression Strategy} & \textbf{B@4} & \textbf{CIDEr} & \textbf{C-S} & \textbf{C-I} & \textbf{Recall}\\
        \midrule
        Fixed & \textbf{22.9} & \textbf{87.5} & \textit{\textbf{12.1}} & \textit{\textbf{8.0}} & \textbf{44.9} \\
        Top-k & 22.8 & 87.3 & 12.2 & 8.1 & 44.6 \\
        Top-(k-1) & 22.6 & 87.1 & 12.3 & 8.1 & 44.6 \\
        Proportional & 22.3 & 87.3 & 12.2 & 8.1 & 44.8 \\
        \bottomrule
    \end{tabular}}
    \caption{Ablation study on selection strategies. Fixed Value Filtering: suppress attention heads with fixed threshold. Top-k/Top-(k-1): select top-k or top-(k-1) attention heads based on the number of negative entities k. Proportional: select top 1\% attention heads proportionally.}
    \label{tab:suppression_strategy}
\end{table}

% \section{Ablation Study on Suppression Strength}
\begin{table}[h]
    \centering
    \scalebox{0.85}{
    \begin{tabular}{c|ccccc}
        \toprule
        \textbf{Suppression Strength ($\lambda$)} & \textbf{B@4} & \textbf{CIDEr} & \textbf{C-S} & \textbf{C-I} & \textbf{Recall}\\
        \midrule
        0.001 & 22.6 & 86.9 & 12.3 & 8.2 & \textbf{45.2} \\
        0.01 & 22.5 & 86.7 & 12.2 & 8.3 & 45.1 \\
        0.1 & 22.7 & 87.2 & \textit{\textbf{12.1}} & 8.2& 45.0\\
        0.3 & \textbf{22.9} & \textbf{87.5} & \textit{\textbf{12.1}} & \textit{\textbf{8.0}} & 44.9 \\
        0.5 & 22.8 & 87.2 & 12.0 & 8.2 & 44.8 \\
        \bottomrule
    \end{tabular}}
    \caption{Ablation study on suppression strength in Flickr30k-to-COCO transfer scenario. $\lambda$ controls the suppression intensity where lower values indicate stronger suppression.}
    \label{tab:suppression_strength}
\end{table}

% \section{Ablation Study on Hallucination Filtering Strategies}
\begin{table}[h]
    \centering
    \scalebox{0.85}{
    \begin{tabular}{l|ccc|ccc}
        \toprule
        \multirow{2}{*}{\textbf{Strategy}} & \multicolumn{3}{c|}{\textbf{Filter Results}} & \multicolumn{3}{c}{\textbf{Captions Hallucination}} \\
        \cmidrule(lr){2-4} \cmidrule(lr){5-7}
        &  Total & RTS & Ratio & CIDEr &C-S & C-I  \\
        \midrule
        % GT & \textit{\textbf{13203}} &\textit{ \textbf{0}} & \textit{\textbf{0 }}& \textbf{90.0} & \textbf{\textit{6.1}} & \textbf{\textit{4.5}} \\
        GT & 13203 & 0 & 0 & 90.0 & 6.1 & 4.5 \\
        \midrule
        NEF & 19811 & 4976 & 25.1 & \textbf{87.5} & \textit{\textbf{12.1}} & 8.0 \\
        CLIP-Det & \textit{\textbf{15369}} & \textit{\textbf{3589}} & \textit{\textbf{23.4}} & 83.8 & 12.3 & \textit{\textbf{7.3}} \\
        \bottomrule
    \end{tabular}}
    \caption{Analysis of hallucination under different filtering strategies. Total: Total hallucinated entities; RTS: Retrieval-sourced hallucinations; Ratio: RTS/Total percentage.}
    \label{tab:nef_analysis}
\end{table}

\section{Detailed Hallucination Analysis}
\label{app:hallucination_analysis}

We extract hallucinated entities from generated captions of ViE, IFCap and our NES model, which we further categorize into \textit{retrieved} (appearing in retrieved captions) and \textit{others}. While ViE lacks retrieval mechanisms (retrieved = 0), IFCap exhibits a concerning pattern: over 20\% of hallucinated entities originate from retrieved content across all tasks. 

We find that while most existing models focus solely on filtering hallucinated information from retrieval results, they overlook the potential for further utilization of this information. We argue that when retrieval models return hallucinated entities, it indicates these entities possess consistent feature representations in the latent space, leading decoders to generate captions containing these retrieved hallucinated entities even without those retrieved information. Rather than merely filtering these entities, our approach recognizes that their presence signals underlying semantic correlations that can be leveraged through targeted suppression mechanisms.

\section{Retrieval Performance Analysis}
\label{app:retrieval_analysis}

As illustrated in Fig.~\ref{fig:intro_retrieval_ana}, we evaluate retrieval performance across different input modalities (original images, synthetic images, and text) using the COCO training set as the retrieval database. We measure entity accuracy (ACC), recall rates (RC), average hallucination counts per image (AHC), and deduplicated hallucination counts (DHC) across multiple CLIP encoder variants. Our analysis yields two key insights:

\textbf{Data Source Perspective:} Original images achieve optimal accuracy and minimal hallucination counts, yet their visual complexity results in poor retrieval recall. Text-based retrieval, while maintaining comparable recall, introduces significantly more hallucinations. Synthetic images strike an optimal balance while reducing hallucination compared to text-based retrieval.

\textbf{Architectural Perspective:} RN50 delivers the highest precision with minimal hallucinations, whereas ViT-L/14 and ViT-B/32, despite superior recall rates, exhibit lower accuracy and higher hallucination rates. Based on these findings, we employ RN50 as the retrieval encoder during the data processing stage to minimize hallucination in retrieved captions, while adopting ViT-B/32 as the feature encoder during model training for its superior compatibility with pretrained models and computational efficiency. This design validates our hypothesis that hallucinated entities from retrieval reflect consistent latent representations that can be effectively suppressed through our proposed attention-level mechanisms during decoding.

\section{Synthetic Image Quality Analysis}
\begin{table}[htbp]
    \centering
    \scalebox{0.85}{
    \begin{tabular}{l|ccc|cc}
        \toprule
        \multirow{2}{*}{\textbf{Dataset}} & \multicolumn{3}{c|}{\textbf{CLIPScore Distribution}} & \multicolumn{2}{c}{\textbf{Quality Thresholds}} \\
        \cmidrule(lr){2-4} \cmidrule(lr){5-6}
        & Mean & Min & Max & High (\%) & Total (\%) \\
        \midrule
        COCO & 81.1 & 40.3 & 93.6 & 67.1 & 87.3 \\
        Flickr30k & 83.2 & 39.9 & 92.4 & 56.4 & 82.6 \\
        \bottomrule
    \end{tabular}}
    \caption{CLIPScore analysis of synthetic images quality.\textit{High: CLIPScore $>$ 0.8; Total: CLIPScore $>$ $\tau$}}
    \label{tab:synthetic_quality}
\end{table}

\section{Image-to-text Retrieval}
\label{app:image-to-text retrieval}
We employ a CLIP-based retrieval framework to obtain semantically relevant captions for input images. Specifically, both the input image and all candidate captions in the datastore are encoded using the CLIP encoder to obtain their respective embeddings in a shared multimodal space. We then perform k-nearest neighbor search based on cosine similarity between the image embedding and text embeddings to retrieve the k most semantically similar captions from the datastore. The impact of varying retrieval sizes (k) on the model's performance is systematically analyzed in Table~\ref{tab:retrieval_num}.

\section{Clip-based Entity Classification}
\label{app:Clip-based Entity Classification}
We implement a zero-shot entity classification approach leveraging CLIP's cross-modal understanding capabilities. First, we construct an entity vocabulary $V$ containing relevant object categories. For each entity $e \in V$, we generate a textual description using the template "A photo of {entity}" to create standardized entity representations. Both the input image and all entity descriptions are encoded using CLIP to obtain their embeddings in the shared feature space. We compute cosine similarities between the image embedding and each entity description embedding, then rank the entities accordingly. Finally, we select the top-M entities with the highest similarity scores as the predicted entity classifications for the input image.

\clearpage
\begin{figure*}[p]
    \centering
    \vspace{-1cm} 
    \includegraphics[width=1\textwidth]{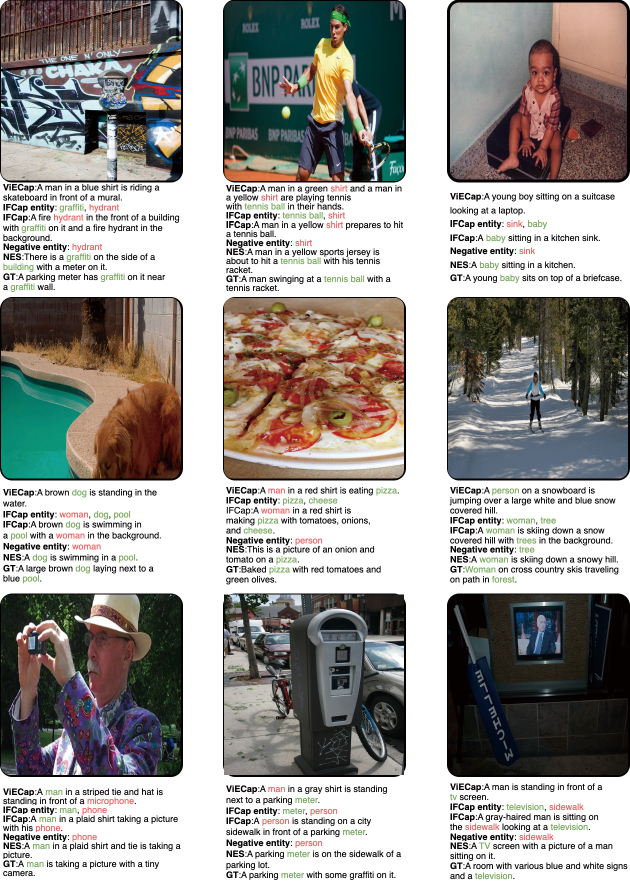}
    \caption{Qualitative result on the Flickr30k-to-COCO test set.}
    \label{fig:qualitative_result}
\end{figure*}

\clearpage

\end{document}